\documentclass[conference]{IEEEtran}

\usepackage{cite}

\usepackage{graphicx}
\usepackage{float}
\usepackage{wrapfig}
\usepackage[normalem]{ulem}
\usepackage{booktabs}

\usepackage{bm}
\usepackage{textcomp}
\usepackage{amssymb}
\usepackage{amsmath}
\usepackage{amsthm}
\usepackage{amsfonts}
\usepackage{mathtools}
\usepackage{pifont}

\usepackage{algorithm}
\usepackage{algpseudocode}

\usepackage{times}
\usepackage{latexsym}
\usepackage{gensymb}

\usepackage{textcomp}
\usepackage{multirow}
\usepackage{lscape}
\usepackage{subcaption}

\usepackage{mdframed}
\usepackage{hyperref}
\usepackage{textgreek}
\usepackage{tabularx, colortbl, xcolor}
\definecolor{mygray}{gray}{0.95}
\definecolor{mycyan}{HTML}{005397}
\definecolor{myred}{HTML}{E13333}
\definecolor{mymagenta}{HTML}{BF3E87}
\definecolor{mypurple}{HTML}{1B2278}

\definecolor{tearose}{HTML}{F584C5}
\definecolor{coral}{HTML}{F67088}
\definecolor{dodger_blue}{HTML}{3BA3EC}
\definecolor{domino}{HTML}{BC9F48}
\definecolor{domino}{HTML}{BC9F48}
\definecolor{catalina_blue}{HTML}{1C3168}
\definecolor{catalina_blue}{HTML}{1C3168}
\definecolor{dark_scarlet}{HTML}{C63D52}
\definecolor{cerulean}{HTML}{0192A8}

\definecolor{tussock}{HTML}{C99E31}
\definecolor{p13}{HTML}{BFB5D7}
\definecolor{b14}{HTML}{BEA1A5}
\definecolor{y15}{HTML}{F0Cf61}
\definecolor{Merino}{HTML}{F3EEE3}

\definecolor{LightOliveGreen}{RGB}{154, 172, 81}
\definecolor{SkyBlue}{RGB}{131, 196, 220}
\definecolor{LightSkyBlue}{RGB}{111, 207, 218}
\definecolor{Lavender}{RGB}{216, 140, 228}

\newcolumntype{a}{>{\columncolor{p13}}l}
\usepackage[capitalize,noabbrev]{cleveref}
\crefname{ineq}{Inequality}{Inequalities}
\creflabelformat{ineq}{#2{\upshape(#1)}#3}

\theoremstyle{remark}
\newtheorem{example}{Example}[]
\usepackage{braket}

\newcommand{\abs}[1]{\left\lvert#1\right\rvert}

\usepackage{stmaryrd}

\DeclareMathOperator*{\argmax}{arg\,max}
\DeclareMathOperator*{\argmin}{arg\,min}

\usepackage{tikz}
\usetikzlibrary{trees}
\usepackage{tikz-dependency}
\usepackage{tikz-3dplot}
\usetikzlibrary{chains,scopes}
\usetikzlibrary{arrows.meta,automata,positioning}
\usetikzlibrary{shapes.geometric, arrows}
\usepackage{pgfplots}

\pgfplotsset{
  every axis/.append style = {thick},
  tick style = {thick,black},
  /tikz/normal shift/.code 2 args = {%
    \pgftransformshift{%
        \pgfpointscale{#2}{\pgfplotspointouternormalvectorofticklabelaxis{#1}}%
    }%
  },%
  shift/.style = {
    tick align        = outside,
    scaled ticks      = false,
    enlargelimits     = false,
    ticklabel shift   = {#1},
    axis lines*       = left,
    xtick style       = {normal shift={x}{#1}},
    ytick style       = {normal shift={y}{#1}},
    x axis line style = {normal shift={x}{#1}},
    y axis line style = {normal shift={y}{#1}},
  },
  shift/.default = 10pt,
  shift3d/.style = {
    shift=#1,
    ztick style       = {normal shift={z}{#1}},
    z axis line style = {normal shift={z}{#1}},
  },
  shift3d/.default = 10pt,
}

\tikzstyle{startstop} = [rectangle, rounded corners, 
minimum width=3cm, 
minimum height=1cm,
text centered, 
draw=black, 
fill=LightOliveGreen!30]

\tikzstyle{io} = [trapezium, 
trapezium stretches=true, % A later addition
trapezium left angle=70, 
trapezium right angle=110, 
minimum width=3cm, 
minimum height=1cm, text centered, 
text width=2.5cm,
draw=black, fill=SkyBlue!30]

\tikzstyle{process} = [rectangle, 
minimum width=3cm, 
minimum height=1cm, 
text centered, 
text width=2.5cm, 
draw=black, 
fill=Lavender!30]

\tikzstyle{example} = [rectangle, 
minimum width=3cm, 
minimum height=1cm, 
text width=4cm,
draw=black, 
fill=LightSkyBlue!30,
dashed,
dash pattern=on 3pt off 3pt]

\tikzstyle{arrow} = [thick,->,>=stealth]

\usepackage{listings}
\usepackage{array}
\newcolumntype{H}{>{\setbox0=\hbox\bgroup}c<{\egroup}@{}}

\begin{document}
\bstctlcite{BSTcontrol}

\title{Label-Free Topic-Focused Summarization Using Query Augmentation
}

\author{
\IEEEauthorblockN{Wenchuan Mu}
\IEEEauthorblockA{\textit{Information Systems Technology and Design} \\
\textit{Singapore University of Technology and Design}\\
Singapore \\
wenchuan\_mu@sutd.edu.sg}
\and
\IEEEauthorblockN{Kwan Hui Lim}
\IEEEauthorblockA{\textit{Information Systems Technology and Design} \\
\textit{Singapore University of Technology and Design}\\
Singapore \\
kwanhui\_lim@sutd.edu.sg}
}

\maketitle

\begin{abstract}
In today's data and information-rich world, summarization techniques are essential in harnessing vast text to extract key information and enhance decision-making and efficiency. In particular, topic-focused summarization is important due to its ability to tailor content to specific aspects of an extended text. However, this usually requires extensive labelled datasets and considerable computational power. This study introduces a novel method, Augmented-Query Summarization (AQS), for topic-focused summarization without the need for extensive labelled datasets, leveraging query augmentation and hierarchical clustering. This approach facilitates the transferability of machine learning models to the task of summarization, circumventing the need for topic-specific training. Through real-world tests, our method demonstrates the ability to generate relevant and accurate summaries, showing its potential as a cost-effective solution in data-rich environments. This innovation paves the way for broader application and accessibility in the field of topic-focused summarization technology, offering a scalable, efficient method for personalized content extraction.


\end{abstract}

\section{Introduction}
In today's world, where we are constantly bombarded with vast amounts of text, the ability to efficiently summarize information has become crucial~\cite{rush-etal-2015-neural}. Summarization, the process of condensing extensive texts into shorter, digestible formats, is important for both individuals and organizations. It enables quicker understanding and better management of information. However, this process is not without its challenges. One major difficulty is creating summaries that are useful for different purposes and to different people, as everyone has unique needs and interests~\cite{bahrainian-etal-2022-newts}.

Addressing these issues, topic-focused summarization offers a new approach. It focuses on generating summaries based on specific topics, making the information more relevant to the reader's interests~\cite{conroy-etal-2006-topic}. This task involves analyzing the content to identify key themes and then highlighting these in the summary. Yet, implementing topic-focused summarization is challenging. It typically requires a lot of organized data to start with and demands high computational power to process this information effectively~\cite{vig-etal-2022-exploring}.

\subsection{Recent Developments and Limitations}
Recent advances in automated summarization have primarily focused on abstractive and extractive techniques~\cite{chu2019meansum}. Abstractive summarization models~\cite{rush-etal-2015-neural}, such as those employing neural language models, attempt to generate a concise and coherent summary by understanding and paraphrasing the original text. These models, often based on complex architectures like transformers and attention mechanisms, have shown remarkable capability in mimicking human-like summary generation~\cite{chu2019meansum,li-etal-2017-cascaded}. However, they are typically dependent on extensive supervised training, requiring large labelled datasets that are not always feasible or available for specific topics or domains~\cite{yu2022survey}. Furthermore, the unconstrained nature of abstractive summarization often leads to challenges in controlling the focus of the summaries, particularly in aligning with specific user queries or topics~\cite{bahrainian-etal-2022-newts}.

\begin{figure*}[t]
    \centering
    \resizebox{0.7\textwidth}{!}{
    \begin{tikzpicture}[node distance=2cm]
    
    \node (start) [startstop] {Start};
    \node (inp) [io, below of=start] {Raw input / Context};
    \node (query) [io, left of=inp, xshift=-1.5cm, yshift=1.8cm] {Topical query};
    \node (ex1) [example, right of=inp, yshift=1.8cm, xshift = 2.2cm] {... I have attached images for reference. Leaking pipes/ceiling in the master bedroom.? need help to resolve asap; unsatisfactory workmanship - this has gone on for multiple times. I have invested time and no one is here to ensure quality. I. ...};
    \node (ex2) [example, left of=query, text width=2.5cm, yshift=0.5cm, xshift = -1cm] {What is the problem?};
    
    \node (pg) [process, below of=query] {Paraphrase generation};
    \node (aq) [io, below of=pg, yshift=0.15cm] {Paraphrased
    topical query};
    \node (ex3) [example, left of=aq, text width=2.4cm, yshift=0.8cm, xshift = -1.35cm] {- What is the problem?
    
    - What is complained?};
    
    \node (qa) [process, below of=inp, yshift=0.15cm] {Question answering};
    \node (ans) [io, right of=qa, xshift=1.5cm] {Answers};
    
    \node (ex4) [example, below of=ans, text width=3cm, yshift=0.5cm, xshift = 3.5cm] {- Leaking pipes/ceiling in the master bedroom. - unsatisfactory workmanship};

    \node (hc) [process, below of=ans] {Hierarchical clustering};
    \node (rans) [io, left of=hc, xshift=-1.5cm] {Relevant answers};
    
    \node (ex5) [example, above of=rans, text width=2cm, yshift=-0.9cm, xshift = 0cm] {(only relevant answers)};

    \node (summ) [process, below of=pg, yshift=-2cm] {Abstractive summarization};
    \node (out) [io, below of=summ, yshift=0.15cm] {Topic focused summary};
    \node (ex6) [example, left of=out, text width=3cm, yshift=0.8cm, xshift = -1.35cm] {Leading pipes/ceiling in master bedroom and unsatisfactory workmanship};
    \node (end) [startstop, right of=out, xshift=1.5cm] {End};

    \draw [arrow] (start) -- (inp);
    \draw [arrow] (start) -- (query);
    \draw [arrow] (query) -- (pg);
    \draw [arrow] (pg) -- (aq);
    \draw [arrow] (aq) -- (qa);
    \draw [arrow] (inp) -- (qa);
    
    \draw [arrow] (qa) -- (ans);
    \draw [arrow] (ans) -- (hc);
    \draw [arrow] (hc) -- (rans);
    
    \draw [arrow] (rans) -- (summ);
    \draw [arrow] (summ) -- (out);
    \draw [arrow] (out) -- (end);
    
    \end{tikzpicture}
  }

    \caption{The Augmented-Query Summarization (AQS) pipeline consists of four pretrained key components: a paraphrasing model, a question-answering model, hierarchical clustering, and an abstractive summarization model. AQS takes two text inputs: the query related to the topic and the typically longer context. It generates a single, topic-focused summary as its output. AQS is an adaptation approach, as all the key components can be derived from generic tasks, such as generic abstractive summarization.}
    \label{fig:aqs}
\end{figure*}
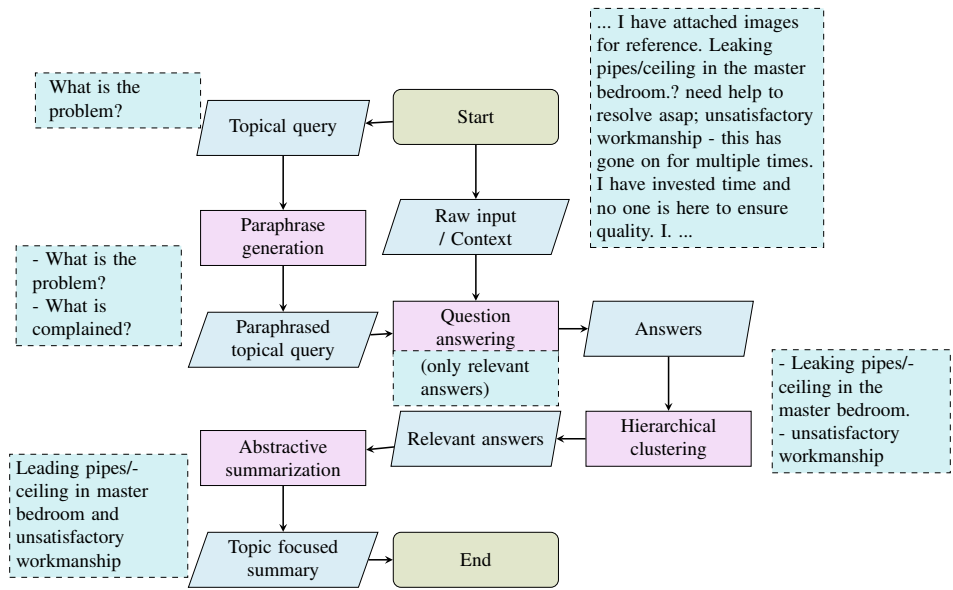

Extractive summarization, on the other hand, involves selecting relevant sentences or segments directly from the source text~\cite{mao-etal-2022-dyle}. This approach, while simpler and more interpretable, often struggles with generating summaries that are as coherent and fluent as those produced by abstractive methods~\cite{see-etal-2017-get,dong2019unified}. Additionally, both abstractive and extractive methods face challenges in topic-focused summarization. They either require substantial customization and retraining for each new topic or fail to adequately prioritize information relevant to specific user queries~\cite{conroy-etal-2006-topic}. This limitation becomes particularly evident in dynamic and diverse real-world applications where the ability to adapt to varying topics and queries rapidly is essential~\cite{vig-etal-2022-exploring}. Therefore, there is a pressing need for a summarization approach that is both flexible and efficient in handling diverse topics without extensive retraining or reliance on large labelled datasets~\cite{yu2022survey}.

\subsection{Our Contributions}
To address the challenges of topic-focused summarization in the absence of labelled data, our proposed method, Augmented-Query Summarization (AQS), employs a novel approach that leverages the strengths of existing natural language processing techniques~\cite{lewis-etal-2020-bart}. AQS integrates four key components: paraphrasing for query augmentation, standard question answering, hierarchical clustering, and generic abstractive summarization. This combination of techniques allows us to effectively generate topic-focused summaries by processing and analyzing the context and content of source texts through a sequence of finely tuned steps. Each component of AQS plays a crucial role in ensuring that the final summary is not only contextually relevant but also concise and coherent, addressing the specific topic of interest without requiring extensive labelled datasets for training.

The practicality and efficiency of AQS are further bolstered by its design, which accommodates the integration of a range of pre-trained models~\cite{lewis-etal-2020-bart,liu2019roberta}. This flexibility not only makes our method adaptable to various domains and topics, but also significantly reduces the time and resources typically required for model training and fine-tuning. Our approach is especially advantageous in situations where rapid deployment and adaptability are essential. Furthermore, by harnessing the capabilities of pre-existing models, AQS stands at the forefront of summarization technology, offering a robust, scalable solution that promises significant improvements in personalized content extraction and information synthesis in data-rich environments. A summary of our key contributions are:

1) Investigating how query and context variations affect the transferability of question-answering models, as well as the impact of input changes on generic abstractive summarization models.
2) Introducing an algorithm that does not require training on topic-focused summarization or extensive labelled datasets.
3) Illustrating the effectiveness and efficiency of our method on real-world data through qualitative and quantitative analysis and experiments.

\section{Related Work}

Abstractive summarization models strive to produce brief, precise, and easily understandable text that captures the most essential information from a document.
In recent years, notable progress has been achieved in the field of generic abstractive summarization~\cite{see-etal-2017-get,gehrmann-etal-2018-bottom,dong2019unified}. This success can be credited to the development of advanced neural architectures and the accessibility of extensive datasets~\cite{rush-etal-2015-neural,vaswani2017attention,devlin-etal-2019-bert}. The BART architecture~\cite{lewis-etal-2020-bart}, for example, employs transformers and implements sequence-to-sequence functionality, eliminating the need for a separate encoder-decoder structure. 

Topic-focused summarization is a more challenging task that involves generating a summary specifically tailored to a given query and its relevant document(s), as illustrated in \cref{ex:tfs}. \cite{mao-etal-2022-dyle} involved treating the extractive text of an extractive-abstractive model as a latent variable. \cite{nema-etal-2017-diversity} introduced an encode-attend-decode system incorporating query and diversity-based attention mechanisms. This approach aimed to produce a summary that is more relevant to the given query. \cite{vig-etal-2022-exploring} trains a relevance prediction model directly on data using the original, non-masked query. \cite{xu-lapata-2021-generating} ranks the sentences in context (specific body of text from which answers to the augmented queries are to be extracted and summarized) according to embedding similarity to create weak-supervising labels before end-to-end training. \cite{su-etal-2022-improving} propose a model that incorporates the explicit answer relevance of the source documents given the query via a question-answering model, to generate coherently and answer-related summaries. \cite{xu-lapata-2020-coarse} and \cite{su-etal-2020-caire} also employed QA models to rank answer evidence at the sentence or paragraph level.  However, these methods might not be able to fully harness the potential of QA models in actively retrieving highly relevant text segments from the context.

Our work leverages the power of pretrained QA models, which excel at attending to queries, with the capabilities of pretrained abstractive summarization models, which generate coherent sentences. In addition, our method is able to combine with any neuron architecture without the training needed.  

\vspace{2mm}

\begin{example}
\label{ex:tfs}

\textbf{Context} $\rhd$ After 30 years of being `stuck' on the spot, the world's biggest iceberg is on the move.  Called A23a, the block of ice is around 1,540 sq miles in area – more than twice the size of Greater London ...\footnote{From Dailymail article ``The world's biggest iceberg is on the move'' \url{https://www.dailymail.co.uk/sciencetech/article-12786619/}}.
\vspace{2mm}

\noindent\underline{Generic summary} $\rhd$ \textit{With an area of 1,540 sq miles, iceberg A23a is the current world record holder
The largest iceberg in the world was A76 before it fragmented into three pieces }
\vspace{1mm}

\noindent\underline{Topical query} $\rhd$\textit{What did Dr Fleming say?}
\vspace{1mm}

\noindent\underline{Topic-focused summary} $\rhd$ A structure, likely an ice shelf, stationary since 1986, began moving due to size reduction over time, with consensus suggesting natural progression rather than temperature changes as the cause.
\end{example}
\vspace{2mm}

\section{Proposed Method: Augmented-Query Summarization (AQS)}

We present Augmented-Query Summarization (AQS), which integrates four elements: paraphrase generation, question answering, hierarchical clustering, and generic abstractive summarization. The workflow of this pipelined method can be viewed in \cref{fig:aqs}.

In this section, we introduce how the proposed pipelined method is constructed and explain how all elements collectively offer a potential solution for topic-focused summarization without training on dedicated data. 

\vspace{2mm}
\subsubsection{Paraphrase generation}

Modifications need to be made to the traditional QA pipeline. In other words, the answers should encompass more than a single segment, allowing for the consideration of information from different parts of the context to provide comprehensive responses. To this end, we resort to tweaking the question/query. The intuition is that when the query is subtly changed while retaining the main focus, we may be able to capture varied content segments. This is because each segment related to the topic can be a valid QA model output, but typically only one is produced. By introducing a degree of uncertainty or variability to the input, we anticipate a broader range of outputs. To this end, our options include varying the query, context, or both. In this study, we primarily delve into modifying the query.

We propose to augment the query through paraphrase generation. By generating multiple queries with the same meaning, we naturally generate multiple potential answers, which may originate from different locations within the context. Additionally, by perturbing the query with semantic constraints, we hypothesize that some of the query paraphrases will effectively activate the transferability of the QA model. The effect of inputting semantically similar but syntactically different queries into the QA model can be captured in Example~\ref{ex:amazon} below.

\vspace{2mm}
\begin{example}
\label{ex:amazon}

\textbf{Context} $\rhd$ The Amazon rainforest includes territory belonging to nine nations. The majority of the forest is contained within Brazil, with 60\% of the rainforest, followed by Peru with 13\%, Colombia with 10\%, and minor amounts in Venezuela, Ecuador, Bolivia, Guyana, ...

\vspace{2mm}
\noindent\underline{Query 1} $\rhd$ \textit{Which country is the 3rd largest in the forest?}
\vspace{1mm}

\noindent\underline{Response} $\rhd$  \textcolor{red}{Peru (incorrect)}
\vspace{1mm}

\noindent\underline{Query 2} $\rhd$\textit{ Which country has the third largest land in the forest?}
\vspace{1mm}

\noindent\underline{Response} $\rhd$ \textcolor{green}{Colombia (correct)}
\end{example}
\vspace{3mm}

We note that a slight syntax change in the query leads to a significant improvement in extracting the correct answer. Although the original and paraphrased queries convey the same meaning, they differ in their phrasing. The improved performance can be attributed to the QA model's sensitivity to query syntax and structure. This phenomenon has also been discussed in literature about the robustness of models~\cite{garg-ramakrishnan-2020-bae}.

The paraphrased query aligns with the patterns and structures learned by the model during training, enabling more accurate identification of relevant information in the context and producing a more precise answer. Therefore, by using paraphrase generation to augment the original query, we can enhance the chances of extracting a satisfactory answer from the context. This part of the method is shown in lines 2-12 in \cref{alg:summ}. Now that we have augmented the topic-related query, the next step is to extract answers using these queries.

\vspace{2mm}
\subsubsection{Question answering}
To address this challenge, we propose to leverage question-answering (QA) models. The intuition of using QA for topic-focused summarization is that we can use the topic to form a question and then extract the topic-related content from the given context. In an adaptation setup, QA is suitable for extracting topic-focused content from a context for the following reasons. First, QA and topic-focused summarization share a common input format, consisting of two texts: one for the query/question and the other for the context. Secondly, they have a comparable output format, both being text strings.

\begin{figure*}[t]
  \centering
  \begin{subfigure}[t]{0.3\linewidth}
    \includegraphics[width=0.97\linewidth, trim={1cm 0 1cm 0},clip]{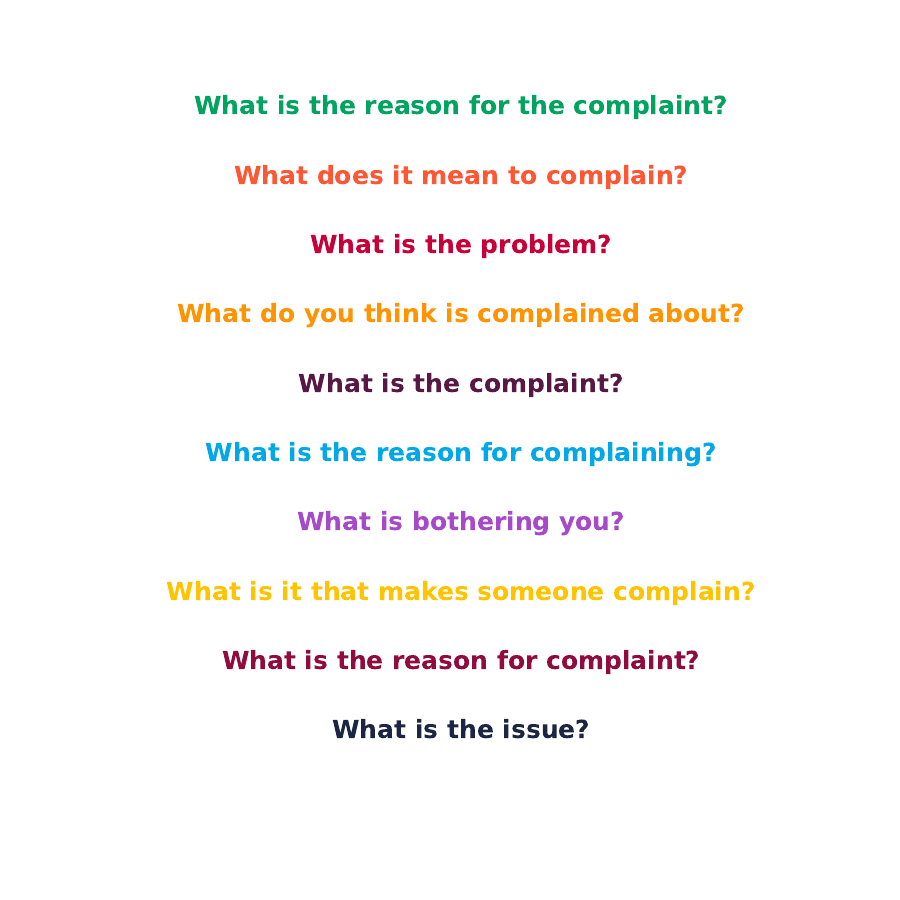}
    \caption{Examples of paraphrased queries. These queries are paraphrased from an original query, ``What is complained?'' }
  \end{subfigure}
  \begin{subfigure}[t]{0.35\linewidth}
    \includegraphics[width=\linewidth]{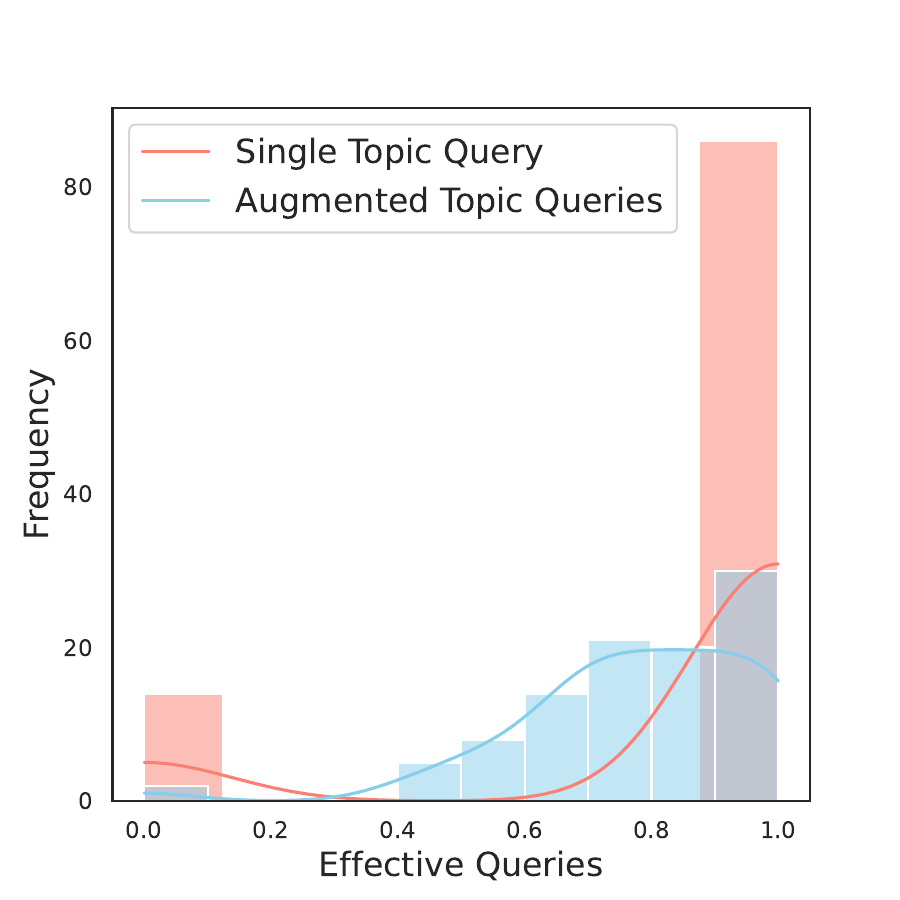}
    \caption{Distribution of the effective queries: Single queries lead to binary outcomes (`correct'/`incorrect'). Multiple queries produce probabilistic outcomes. Horizontal axis: the proportion of queries with correct answers (effective queries). Vertical axis: the number of documents input (\%). Line curves: kernel density estimation.}
  \end{subfigure}
  \begin{subfigure}[t]{0.325\linewidth}
    \raisebox{4mm}{\includegraphics[width=\linewidth]{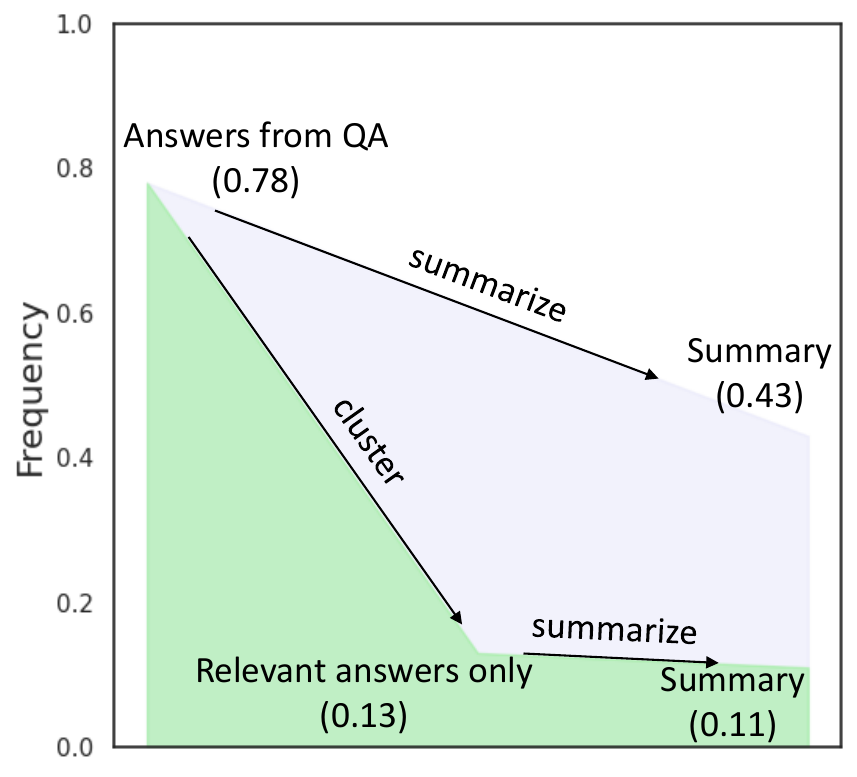}}
    \caption{Text redundancy, a measure of irrelevant content in answers or summarised answers. Without clustering, answer redundancy is 0.78, leading to a summary redundancy of 0.43. With clustering, answer redundancy decreases to 0.13, resulting in a summary redundancy of 0.11. }
  \end{subfigure}
  
  \caption{Illustration of the functionality of each part in the proposed method. Paraphrasing generates multiple queries from a single one. Using multiple queries may stabilise the QA model performance. On some inputs, a single query fails to yield a correct answer, while using multiple queries may yield, say 70\%, correct answers.  When the majority part of answers are correct, then the answer can be considered as correct. Note that a correct answer from QA might also contain irrelevant content, \emph{e.g.}, function words.  Clustering significantly reduces redundancy in QA. This is likely because some incorrect answers are pretty long, and thus account for a large proportion of content. Therefore, removing these incorrect answers before summarization is needed. Besides, summarization helps reform a fluent sentence from the clustering-selected answers. Intuitively, the text redundancy in the summaries is even lower.}
  \label{fig:illus}
\end{figure*}

However, there are also challenges in using QA for topic-focused summarization, as there is a key distinction in their output constraints. In QA, the output is typically a substring extracted from the context, whereas topic-focused summarization does not impose such limitations. In other words, the (single-answer) QA task is a task of extracting a single segment from the context, whereas topic-focused summarization may expect content from many discrete places in the context. We suggest a parallel question-answering method that processes augmented queries alongside a singular context. In this approach, every query is matched with the context to generate a response. Theoretically, if a standard QA model has a probability $p$ of producing successful answers, the efficacy of the collective response improves as the quantity of augmented queries rises, converging towards $p$. Moreover, if $p$ exceeds 1/2, it is highly probable to obtain a successful answer, as depicted in \cref{fig:illus}(b).

\vspace{2mm}
\subsubsection{Hierarchical clustering}
The challenge that comes along with using an augmented set of queries is effectively managing the increased complexity and diversity of the query space. When we introduce a larger set of augmented queries, it becomes crucial to ensure that the QA system can efficiently process and evaluate each query in a timely manner. The system should be able to handle the expanded query space without sacrificing performance or introducing unnecessary computational overhead. We need to determine the optimal balance between query diversity and redundancy. While generating a diverse set of augmented queries can enhance the chances of capturing different perspectives and nuances, it also increases the likelihood of introducing redundant or overlapping queries. Managing this redundancy becomes important to avoid excessive computational costs and potential confusion in the answer selection process.

Moreover, there is a need to assess the quality and relevance of the augmented queries. Not all generated paraphrases or variations may be equally useful or effective in extracting the desired information from the context. Careful evaluation and filtering techniques are necessary to identify the most promising queries that are likely to yield satisfactory answers.

Addressing these challenges requires the development of efficient query management strategies, intelligent query selection techniques, and robust evaluation mechanisms. By effectively addressing these challenges, we can harness the benefits of using an augmented set of queries to improve the overall performance and effectiveness of QA systems. Thus, we propose clustering answers into relevant and irrelevant ones, eliminating off-topic or poorly formatted answers, to improve coherence and focus in the summarized output.

Hierarchical clustering is employed in this approach, where a symmetric similarity score is calculated between each pair of answers. The two most similar answers are then combined into a longer answer. This process is repeated until the longest answer contains at least half of the original individual answers. Subsequently, the longest answer is used as input for the abstractive summarization model. Refer to lines 17-20 in \cref{alg:summ} for a detailed description.

\begin{algorithm*}[t]
\caption{Topic-Focused Summarization  $(\mathbf{x}, \mathbf{z}, n\in \mathbb{N}_+), q\in[0, 1)$}
\label{alg:summ}
\begin{algorithmic}[1]
\Require{$\mathbf{x}$: Topic inquiry, $\mathbf{z}$: Input text, $n$: Beam size, $q$: patience factor}
\State{$\mathcal{H}_{cur}\gets \{(\epsilon, 0.0)\}$} \Comment{Initialize with empty generation prefix and zero score}
\Repeat
  \State{$\mathcal{H}_{next}\gets \emptyset$}
  \For{$(\mathbf{y},p)\in \mathcal{H}_{cur}$}
    \If{$y_{|\mathbf{y}|}=\mathtt{</s>}$} 
      \State{$\mathcal{H}_{next}\gets \mathcal{H}_{next} \cup \{(\mathbf{y},p)\}$}
      \Comment{Hypotheses ending with $\mathtt{</s>}$ are not expanded}
    \Else
      \State{$\mathcal{H}_{next}\gets \mathcal{H}_{next} \cup \bigcup_{w\in \mathcal{T}} (\mathbf{y}\cdot w, p + \log P(w|\mathbf{x},\mathbf{y}))$}
      \Comment{Add all possible continuations}
    \EndIf
  \EndFor
  \State{$\mathcal{H}_{cur}\gets \{(\mathbf{y},p)\in \mathcal{H}_{next} : |\{(\mathbf{y}',p')\in \mathcal{H}_{next} : p' > p\}| < n\}$}
  \Comment{Select $n$-best}
\Until{$\forall (\mathbf{y},p)\in \mathcal{H}_{cur}.~ y = \mathtt{</s>}$}
\State{$\mathcal{H}_{seg}\gets\emptyset$}
\For{$(\mathbf{y},p)\in \mathcal{H}_{cur}$}
\State{$\mathcal{H}_{seg}\gets \mathcal{H}_{seg} \cup \set{Q(\mathbf{y}\cdot \mathbf{z})}$} \Comment{Each answer is a group}
\EndFor
\Repeat
\State{$(\mathbf{u}, \mathbf{v})\gets \argmin_{i, j\in \mathcal{H}_{seg}} d(i, j)$ }
\State{$\mathcal{H}_{seg}\gets \mathcal{H}_{seg} \cup \set{\mathbf{u} \cdot \mathbf{v}} \setminus \set{\mathbf{u}, \mathbf{v}}$ } \Comment{Merge the two most similar groups}
\Until{$\max_{\mathbf{u}\in \mathcal{H}_{seg}} \abs{\mathbf{u}} > q \sum_{\mathbf{u}\in \mathcal{H}_{seg}} \abs{\mathbf{u}}$}
\Return{$S(\argmax_{\mathbf{u}\in \mathcal{H}_{seg}} \abs{\mathbf{u}}) $} \Comment{Summarize the largest group}
\end{algorithmic}
\end{algorithm*}

\vspace{2mm}
\subsubsection{Generic abstractive summarization}
Now that we have obtained effective answers, the next step is to write a fluent overall answer. We hypothesize that a generic text-to-text model can cope with this problem. Generic abstractive summarization condenses text by generating concise summaries that capture main ideas and context. Moreover, there are abundant off-the-shelf models. The following example illustrates the impact of introducing irrelevant information into a text on the performance of abstractive summarization models. Such perturbations can lead to a decrease in the quality of the generated summaries.

\vspace{2mm}

\begin{example}
\label{ex:others}

\noindent\underline{Input} $\rhd$ \textit{Brief of Incident, Subject Vehicle Illegal passing the gantry place1Case 01, Gantry at place1Case 01 Vehicle Illegal.}

\vspace{1mm}
\noindent\underline{Generated Summary} $\rhd$ \textcolor{green}{Case 01, Gantry at place1. Vehicle Illegal passing the gantry. (improved)}

\vspace{2mm}
\noindent\underline{Slightly Changed Input} $\rhd$ \textit{Thank you subject vehicle ID 4e9aM0\$3N5Thank you Incident, Subject Vehicle Illegal passing the gantry placeCase 01, Gantry at place1Case 01 Vehicle Illegal, Thank you.}

\vspace{1mm}
\noindent\underline{Generated Summary} $\rhd$ \textcolor{red}{Vehicle ID 4e9aM0\$3N5Thank you Incident, (hardly readable)}
\end{example}
\vspace{2mm}

However, current abstractive summarizers, especially neural models, can be sensitive to input syntax, resulting in potential errors or inconsistencies (Example~\ref{ex:others}). It is unwise to directly input answers from QA tasks into abstractive summarizers for two reasons. Firstly, answers with bad syntax or characters extracted from the original context can cause out-of-distribution issues. Secondly, redundancy in answers can propagate to the summaries if concatenated directly. Therefore, we need to use hierarchical clustering to select appropriate queries among all we obtain from paraphrase generation, which leads us to the topic-focused summary.

\section{Experiment and Result}
\begin{table*}[t]
    \centering
    \small
    \caption{Estate Complaints/Feedback (ECF) Dataset Demo with five fields: Subject, Description, Category, Sub Category, and Record Key. It includes 13 categories and 64 subcategories. The Subject field often lacks or is poorly written, necessitating topic-focused summarization.}
    \label{ex:osa}
    \begin{tabular}{|p{1.85cm}|p{2.85cm}|p{1.85cm}|p{2.85cm}|p{5.8cm}|}
    \hline
    \rowcolor{mygray} 
    \textbf{Unique Case Record Key} & \textbf{Subject} & \textbf{Reporting Category} & \textbf{Reporting Sub Category} & \textbf{Description} \\
    \hline
    1***81 & FW: Vehicle Illegal Bypassing the Gantry at *place*-Case 01 & Illegal Parking & Illegal Parking - Public *org* Car Parks/Service Roads & \scriptsize *date*  *time* PM To ...
    Subject Vehicle Illegal Bypassing the Gantry at *place* 01 á Dear *name* á Brief of Incident á On the ... á á Please see the document as attached á á Thank You á Best Regards ... *org* Pte Ltd *org* \\
    \hline
    \end{tabular}
\end{table*}

\begin{table*}[t]
\vspace{3mm}
\centering
\caption{ROUGE-F1 scores are presented for the QMSum and Debatepedia datasets. The previous work can be classified into three categories: 1) pre-trained transformers on a generic summarization dataset, 2) fine-tuning of these transformers on a query-focused summarization dataset, and 3) models specifically designed for supervised training in query-focused summarization.}
\label{tab:qfs}
\small
\scalebox{0.85}{
    \begin{tabular}{m{0.2\textwidth}|*{3}{>{\centering\arraybackslash}p{0.06\textwidth}}|*{3}{>{\centering\arraybackslash}p{0.06\textwidth}}}
    \toprule
    Approach (Train set) & \multicolumn{3}{c|}{Validate on QMSum} & \multicolumn{3}{c}{Validate on Debatepedia} \\
    \cline{2-7}
    & R-1 & R-2 & R-L & R-1 & R-2 & R-L\\
    \midrule
    BART (CNN/Dailymail) & 31.87 & 9.08 & 27.50 & 37.03 & 27.63 & 36.77 \\
    T5 (CNN/Dailymail) & 32.45 & \textbf{9.80} & 28.48 & 39.72 & 29.33 & 37.37  \\
    CAiRE-COVID & 30.98 & 4.70 & 25.80 & 13.32 & 2.47 &  12.18 \\
    BART (QMSum) & \cellcolor{gray!25}\textcolor{gray}{32.42} & \cellcolor{gray!25}\textcolor{gray}{9.62} & \cellcolor{gray!25}\textcolor{gray}{28.37} & \textit{42.11} & \textbf{38.04} & \textit{41.42} \\
    T5 (QMSum) & \cellcolor{gray!25}\textcolor{gray}{34.24} & \cellcolor{gray!25}\textcolor{gray}{9.30} & \cellcolor{gray!25}\textcolor{gray}{28.91} & 40.96 & 37.61 & 38.71 \\
    SEGENC (QMSum) & \cellcolor{gray!25}\textcolor{gray}{37.80} & \cellcolor{gray!25}\textcolor{gray}{13.43} & \cellcolor{gray!25}\textcolor{gray}{33.38} & \textit{41.50} & \textit{38.02} & \textit{39.83} \\
    QFS-BART (QMSum) & \cellcolor{gray!25}\textcolor{gray}{36.07} & \cellcolor{gray!25}\textcolor{gray}{10.89} & \cellcolor{gray!25}\textcolor{gray}{34.65} & 38.94 & 36.68 & 39.63 \\
    BART (Debatepedia) & 31.93 & 8.33 & 27.27 & \cellcolor{gray!25}\textcolor{gray}{57.96} & \cellcolor{gray!25}\textcolor{gray}{44.09} & \cellcolor{gray!25}\textcolor{gray}{57.40} \\
    T5 (Debatepedia) & 31.38 & 8.39 & 27.85 & \cellcolor{gray!25}\textcolor{gray}{53.45} & \cellcolor{gray!25}\textcolor{gray}{40.39} & \cellcolor{gray!25}\textcolor{gray}{52.42} \\
    SEGENC (Debatepedia) & \textit{33.77} & \textit{9.51} & \textit{31.86} & \cellcolor{gray!25}\textcolor{gray}{54.41} & \cellcolor{gray!25}\textcolor{gray}{38.49} & \cellcolor{gray!25}\textcolor{gray}{51.18} \\
    QFS-BART (Debatepedia) & \textit{33.30} & 8.94 & \textit{30.76} & \cellcolor{gray!25}\textcolor{gray}{59.02} & \cellcolor{gray!25}\textcolor{gray}{44.59} & \cellcolor{gray!25}\textcolor{gray}{57.44} \\
    AQS (\textbf{Label-free}) & \textbf{33.84} & \textit{9.79} & \textbf{32.38} & \textbf{46.07} & \textit{37.98} & \textbf{44.43} \\
    
    \midrule
    \rowcolor{mygray}
    \multicolumn{7}{p{0.7\linewidth}}{
         The best adaptation method is indicated in \textbf{bold}, second and third \textit{italics}, and shaded cells represent different splits of the same dataset, not demonstrating adaptation behaviour.}\\
    \bottomrule
    \end{tabular}
    }
\end{table*}

\begin{table*}[t]
\centering
\caption{ROUGE-F1 scores and Bert-Score are reported for the real-world ECF dataset. Moreover, we present the sentiment consistency (CNS) and human satisfactory score, as described in \cref{sec:setup}. All the methods listed are applied in adaptation mode, \emph{i.e.}, with no training or fine-tuning on the tested distribution.}
\small
\begin{tabular}{m{0.2\textwidth}|*{3}{>{\centering\arraybackslash}p{0.06\textwidth}}|*{3}{>{\centering\arraybackslash}p{0.06\textwidth}}}
\toprule
Approach & R-1 & R-2 & R-L & Bert-Score & Sentiment CNS. & Human Eval\\
\midrule
BART (CNN/Dailymail) & 4.23 & 2.02 & 3.84 & 82.17 & 0.211 & 1.95 \\
T5 (CNN/Dailymail) & 3.51 & 1.47 & 2.33 & 84.62 & 0.219 & 0.63 \\
CAiRE-COVID & 11.97 & 2.23 & 11.78 & 86.56 & 0.672 &  4.03 \\
BART (QMSum) & 10.91 & 3.45 & 8.74 & 82.48 & 0.607 & 2.48 \\
T5 (QMSum) & 12.78 & 3.92 & 9.58 & 84.93 & 0.699 & 3.87 \\
SEGENC (QMSum) & 15.87 & 4.25 & \textit{12.64} & \textit{88.56} & 0.493  & \textit{4.39} \\
QFS-BART (QMSum) & 13.03 & 4.29 & 12.61 & 87.88 & 0.515  & 2.01 \\
BART (Debatepedia) & 2.71 & 0.63 & 1.17 & 86.47 & 0.528 & 0.42 \\
T5 (Debatepedia) & 4.35 & 1.04 & 2.98 & 81.22 & 0.716 & 3.57 \\
SEGENC (Debatepedia) & \textit{16.23} & \textit{5.92} & 11.09 & 86.79 & \textit{0.745}  & 4.15 \\
QFS-BART (Debatepedia) & 10.56 & 4.78 & 9.26 & 82.97 & 0.453 & 1.33 \\
AQS (\textbf{Label-free})  & \textbf{21.94} & \textbf{6.37} & \textbf{15.68} & \textbf{88.75} &\textbf{0.807} &  \textbf{4.82} \\
\midrule
\rowcolor{mygray}\multicolumn{7}{l}{The best adaptation method is indicated in \textbf{bold}, the runner-up in \textit{italics}}\\
\bottomrule
\end{tabular}
\label{tab:osa}
\end{table*}

\begin{table*}[t]
\centering
\caption{Ablation study on the components of the proposed AQS. In rows 2 to 4, we made the following modifications: a) replacing the pretrained transformer from BART to T5, b) removing the clustering step, and c) eliminating the query paraphrasing step. This table shows what are the factors influential/non-influential to our method.}
\small
\begin{tabular}{m{0.2\textwidth}|*{3}{>{\centering\arraybackslash}p{0.06\textwidth}}|*{3}{>{\centering\arraybackslash}p{0.06\textwidth}}}
\toprule
Approach & R-1 & R-2 & R-L & Bert-Score & Sentiment CNS. & Human Eval\\
\midrule
AQS & \textbf{21.94} & \textit{6.37} & \textbf{15.68} & \textbf{88.75} &\textit{0.807} &  \textbf{4.82} \\
AQS - BART + T5 & \textit{19.40} & \textbf{6.38} & \textit{13.29} & \textit{88.60} &\textbf{0.833} &  \textit{4.40} \\
AQS - Clus. & 6.47 & 1.22 & 2.39 & 79.45 & 0.203 & 1.57 \\
AQS - Para. & 1.51 & 0.17 & 1.23 & 81.43 & 0.208 & 0.47 \\
\midrule
\rowcolor{mygray}\multicolumn{7}{l}{The best adaptation method is indicated in \textbf{bold}, the runner-up in \textit{italics}}\\
\bottomrule
\end{tabular}
\label{tab:ablation}
\end{table*}

\subsection{Setup}
\label{sec:setup}

\paragraph{AQS Implementation} ~We implement AQS with 4 established models: Pegasus~\cite{zhang2020pegasus} for paraphrase generation, BART~\cite{lewis-etal-2020-bart} for QA, RoBERTa~\cite{liu2019roberta} for text embedding, and BART for abstractive summarization. These models are respectively pretrained on the QQP dataset \footnote{\url{https://quoradata.quora.com/First-Quora-Dataset-Release-Question-Pairs}}, the SQuAD~\cite{rajpurkar-etal-2016-squad}, and the CNN/Dailymail dataset~\cite{hermann2015teaching}. While specific models are utilized in our experiments, our framework's design allows for the substitution of comparable models, such as T5~\cite{raffel2019exploring}, to affirm its versatility. It is critical to underscore that all models are trained on generic datasets recognized for each task's benchmarking, such as SQuAD for QA and CNN/Dailymail for summarization, without fine-tuning on the same or similar datasets for the task at hand. This ensures the validity of adaptation in the proposed method. The paragraph used is shown in \cref{fig:illus}(a) and they are generated using the off-the-shelf model.

\vspace{2mm}
\paragraph{Validation Datasets} ~We assess AQS on two benchmark datasets: Debatepedia~\cite{nema-etal-2017-diversity} and QMSum~\cite{zhong-etal-2021-QMSum}. Debatepedia comprises 13,719 query-context-summary triplets, specifically designed for debate summarization tasks. QMSum offers a collection of 1,808 triplets from 232 English-language meetings across varied fields, including product design, academia, and politics. AQS is further evaluated on the real-world Estate Complaints/Feedback (ECF) dataset to avoid adaptive overfitting, which contains feedback records for an estate department from 2016 to 2021. The structure of the ECF dataset samples is outlined in \cref{ex:osa}. The evaluation utilizes the most recent data split from 2021, with 161,101 samples. For ECF Feedback, we use two fixed queries: ``What is complained?'' to identify the main issues, and ``What is the emotion?'' to determine sentiment tones.

\vspace{2mm}
\paragraph{Metrics} ~ROUGE~\cite{lin-2004-rouge} is the standard for summarization model assessment. This work utilizes (F1) ROUGE-1, 2, and L for automatic model evaluation on QMSum and Debatepedia. For the ECF dataset, we measure ROUGE and Bert-Score~\cite{zhang2019bertscore} with reference to the Reporting Sub Category. A sentiment analysis model\footnote{Sentiment analysis pipeline from Hugging Face \url{https://huggingface.co/blog/sentiment-analysis-python}} compares original and summary sentiment, using Pearson's correlation. Two postHuman Annotators also conduct manual evaluations on a random 50-sample subset from ECF's 2021 data.

\vspace{2mm}
\paragraph{Baselines} ~We benchmark our AQS against 1) Pretrained transformers like T5~\cite{raffel2019exploring} which are conventionally adapted for topic-focused summarization at inference; 2) Transformers fine-tuned on combined queries and contexts, suitable for supervised or adaptation modes; 3) Other label-free models, such as CAiRE-COVID, originally design for shortlisting COVID-19 information~\cite{su-etal-2020-caire}; 4) Advanced end-to-end methods such as QFS-BART~\cite{su-etal-2021-improve} or SEGENC~\cite{vig-etal-2022-exploring}. This method can also be employed in either supervised or adaptation mode. For instance, SEGENC can be trained on QMSum and then validated on Debatepedia or ECF dataset.

\subsection{Result}

\paragraph{Effectiveness}
Our primary goal is to assess how well a method performs on a new dataset it has not seen during training. This `different dataset' is distinct from merely splitting the training and validation data. As shown in \cref{tab:qfs}, AQS outperforms other methods on both the QMSum and Debatepedia datasets, in terms of summary effectiveness. The summary effectiveness is measured by the closeness~\cite{lin-2004-rouge} between the generated summary and the reference in the test dataset. In contrast, when generic summarization models are used for query-focused summarization, their effectiveness is notably limited. Similarly, dedicated query-focused summarization methods find it challenging to adapt to shifts in data distribution between training and targeted test data. In short, AQS produce well-formed summaries compared to baseline methods on unseen topic-focused summarization tasks.

\vspace{2mm}
\paragraph{Efficiency}

AQS, featuring four transformer models with approximately $380\pm100$ M parameters each, emphasizes efficiency. It can execute sequentially (or parallel), sidestepping processing bottlenecks, and achieves rapid performance, processing 183 samples per minute on a single NVIDIA Tesla T4 GPU. In contrast, QFS-BART requires double the GPU memory due to larger model dimensions, while SEGENC, over ten times larger than AQS, demands a high-end A100 GPU. AQS represents a significant step forward in optimizing the balance between model complexity and operational efficiency.

\vspace{2mm}
\paragraph{Use Case}

Since the ECF dataset lacks reference summaries, the Reporting Sub Category serves as a reference. While exact matches are not always necessary between sub-category names and predicted summaries, assessing semantic similarity can gauge model performance. When a summary closely aligns with its sub-category name, it indicates better performance. Similarly, sentiment preservation evaluation matches the predicted summary with the original sentiment. As shown in \cref{tab:osa}, AQS generates highly favourable summaries for Descriptions, benefiting from its use of well-pretrained transformers with strong transferability from generic task training.

\vspace{2mm}
\paragraph{Ablation}
We explore how AQS effectively combines pre-trained transformers for topic-focused summarization. The ablation study presented in \cref{tab:ablation} examines the performance when gradually removing specific components of AQS. The results reveal that a significant drop in performance occurs if either the clustering or paraphrasing part is excluded. Without clustering, the presence of redundant text from the answers can dilute the relevant information, leading to off-topic summaries. Similarly, without paraphrase generation, relying solely on a QA model results in a single, often short answer, rendering the presence or absence of downstream models inconsequential. \cref{fig:illus} visually demonstrates the impact of AQS's paraphrase module and clustering module on the relevance of information at each stage of text processing.

\vspace{2mm}
\section{Conclusion}

This study investigates how pretrained neural models can be adapted to create summaries focused on specific topics, especially when there are no labelled examples of such summaries available. Our approach involves using paraphase-based query augmentation, question answering, and hierarchical clustering to shift from generic summary generation to topic-specific summarization. This method does not need data specifically labelled for topic-focused summarization. Instead, it uses four distinct elements that can be developed separately and thus benefit from a broader range of data, such as that used for generic abstractive summarization. The effectiveness of this method is validated on an existing test set for topic-focused summarization. Besides, this strategy presents a viable option for real-world applications without demanding substantial computational resources.

\vspace{3mm}

{\small
\noindent\textbf{Acknowledgment.}
This research/project is supported by the National Research Foundation, Singapore, and Ministry of National Development, Singapore under its Cities of Tomorrow R\&D Programme (CoT Award COT-V2-2020-1). Any opinions, findings and conclusions or recommendations expressed in this material are those of the author(s) and do not reflect the views of National Research Foundation, Singapore and Ministry of National Development, Singapore.
K. H. Lim is supported by a MOE AcRF Tier 2 (MOE-T2EP20123-0015).
}

% Generated by IEEEtran.bst, version: 1.14 (2015/08/26)

\end{document}